\documentclass[letterpaper, 10 pt, journal, twoside]{IEEEtran}

\IEEEoverridecommandlockouts                              

\usepackage{hyperref}
\usepackage{graphicx}
\usepackage{amsfonts}
\usepackage[dvipsnames]{xcolor}
\usepackage{amsmath} 
\usepackage{amssymb}
\usepackage[ruled,vlined]{algorithm2e}
\SetKwInput{KwInput}{Input}
\usepackage[export]{adjustbox}
\usepackage{siunitx}
\DeclareSIUnit{\impulseunit}{N.s}
\DeclareSIUnit{\momentunit}{N.m}

\newcommand{\com}{\boldsymbol{c}}
\newcommand{\dcm}{\boldsymbol{\xi}}
\newcommand{\cop}{\boldsymbol{u_0}}
\newcommand{\nextstep}{\boldsymbol{u_T}}
\newcommand{\offset}{\boldsymbol{b_T}}
\newcommand{\gencoor}{\boldsymbol{q}}
\newcommand{\genvel}{\boldsymbol{v}}
\newcommand{\nonlinear}{\boldsymbol{h}}
\newcommand{\nonlinearconstant}{\boldsymbol{h_c}}
\newcommand{\torque}{\boldsymbol{\tau}}
\newcommand{\force}{\boldsymbol{\lambda}}
\newcommand{\foot}{\boldsymbol{x}}
\newcommand{\footforce}{\boldsymbol{f}}
\newcommand{\selection}{\boldsymbol{s}}

\begin{document}

\title{Variable Horizon MPC with Swing Foot Dynamics\\ for Bipedal Walking Control
}

\author{Elham Daneshmand$^{1,2}$, Majid Khadiv$^{1}$, Felix Grimminger$^{1}$, Ludovic Righetti$^{1,3}$
\thanks{Manuscript received: October, 15, 2020; Revised January, 18, 2021; Accepted February, 15, 2021.}
\thanks{This paper was recommended for publication by Editor Abderrahmane Kheddar upon evaluation of the Associate Editor and Reviewers' comments.
This work was supported by the Max-Planck Institute for Intelligent Systems' Grassroots program (M10338), New York University, the European Union’s Horizon 2020 research and innovation program (grant agreement 780684 and European Research Council’s grant 637935) and the National Science Foundation (CMMI-1825993).}
\thanks{$^{1}$Max Planck Institute for Intelligent Systems, Tübingen, Germany
        {\tt\footnotesize firstname.lastname@tuebingen.mpg.de}}%
\thanks{$^{2}$Amirkabir University of Technology, Tehran, Iran
        {\tt\footnotesize elham.daneshmand@aut.ac.ir}}
\thanks{$^{3}$ Tandon School of Engineering, New York University, Brooklyn, USA {\tt \footnotesize ludovic.righetti@nyu.edu}}
\thanks{Digital Object Identifier (DOI): see top of this page.}
}

\newtheorem{remark}{\textbf{Remark}}

\markboth{IEEE Robotics and Automation Letters. Preprint Version. Accepted February, 2021}
{Daneshmand \MakeLowercase{\textit{et al.}}: Variable Horizon MPC with Swing Foot Dynamics  for Bipedal Walking Control} 

\maketitle

\begin{abstract}
In this paper, we present a novel two-level variable Horizon Model Predictive Control (VH-MPC) framework for bipedal locomotion. In this framework, the higher level computes the landing location and timing (horizon length) of the swing foot to stabilize the unstable part of the center of mass (CoM) dynamics, using feedback from the CoM states. The lower level takes into account the swing foot dynamics and generates dynamically consistent trajectories for landing at the desired time as close as possible to the desired location. To do that, we use a simplified model of the robot dynamics projected in swing foot space that takes into account joint torque constraints as well as the friction cone constraints of the stance foot. We show the effectiveness of our proposed control framework by implementing robust walking patterns on our torque-controlled and open-source biped robot, Bolt. We report extensive simulations and real robot experiments in the presence of various disturbances and uncertainties.
\end{abstract}


\section{Introduction}
\IEEEPARstart{H}{umanoid} robots should be able to walk robustly on different terrains in the presence of various uncertainties. Hence, the main goal of a walking controller is to find an optimal set of contact schedule and feasible contact forces to robustly achieve a desired task. However, since the system is hybrid, nonlinear and highly constrained, solving the problem holistically is extremely hard \cite{posa2014direct,lengagne2013generation}. That is why most approaches based on optimal control use simplified dynamic models and solve multi-stage optimization to enable real-time computations. The main paradigm to break down the problem is to first decide the optimal contact sequence \cite{deits2014footstep,tonneau2018efficient,lin2019efficient} and then optimize over the contact forces \cite{dai2016planning,carpentier2018multicontact} sometimes together with step location and timing adaptation \cite{ponton2020efficient}.

For bipedal walking on regular surfaces, there is no ambiguity in contact sequence, i.e. the role of the stance and swing foot switches at each walking phase. Therefore, a walking controller should decide where and when to step and how to control highly constrained interaction forces to generate a desired walking behaviour and reject disturbances. Using the linear inverted pendulum model (LIPM), a walking controller can be written as a linear quadratic program that can be solved quickly \cite{kajita2003biped}. Therefore, Linear Model predictive control (MPC) has become a powerful tool for controlling bipedal walking \cite{wieber2006trajectory,herdt2010online,feng2016robust,khadiv2020walking, jeong2019robust}. The optimal control problem (OCP) solved at each control cycle in these approaches focuses on the center of mass (CoM) dynamics and investigates how to stabilize it using force modulation, for predefined or adaptive step locations and timings. They all consider some proxy constraints to guide roughly the swing foot to touch the ground at the desired time and location. However, in order to make sure that this is the case, one needs to generate a feasible swing foot trajectory that establishes contact at the desired time and location, and be consistent with the CoM trajectory and planned step location and time. 

Establishing contact is equivalent to imposing a state dependent switching constraint at a certain time, i.e. the swing foot should touch the ground at that time. This can be formalized using a variable horizon MPC (VH-MPC) framework \cite{shekhar2012robust,mirzaeva2015harmonic} where the main goal is to establish contact at a certain time (the horizon length which is a decision variable). In the VH-MPC framework, a terminal constraint (and cost) is considered at the end of the horizon, but rather than having a moving horizon with a fixed duration, the horizon length changes such that the switching constraint at a certain time remains a terminal constraint. In the special case where the time of the terminal constraint is fixed, this frameworks becomes similar to a shrinking horizon MPC  \cite{farahani2018shrinking,farooqi2020shrinking}. 

We argue that VH-MPC is a suitable framework to control walking of legged robots for two reasons. First, using VH-MPC, we can stabilize the CoM dynamics by only looking at the end of the current step (switching surface/manifold), optimizing both step location and timing, without a need to consider several steps (switches) in the horizon \cite{khadiv2020walking}. Second, the swing foot should land on the ground at a designed time and this is a final-value problem after which the swing foot becomes stance with different control objectives (modulate force to control the CoM). It is important to note that there is no reason in general to track a swing foot trajectory, the only thing that matters is landing at a certain time at a desired location. That is why VH-MPC \emph{with terminal constraints} is a suitable approach to control the swing foot motion.

In this paper, we propose a two-level VH-MPC framework, where the high-level MPC adapts the next step location and timing to stabilize the divergent component of motion (DCM) of the CoM dynamics, and the low-level MPC takes into account the swing foot dynamics to land as close as possible to the planned location at the desired time. For the swing foot dynamics, we project the dynamics of the robot to the swing foot space. Since this dynamic model is nonlinear, we use a linearized version of it. We also take into account the stance foot friction cone and joint torque constraints to find bounds on the actuation force that can be (virtually) applied by the swing foot. Using the swing foot dynamics and constraint on this actuation force, we write down a VH-MPC problem that finds a dynamically consistent trajectory for the swing foot to land at the desired position at the desired time given by the high-level MPC. Note that in the swing foot MPC problem, the horizon is given at each control cycle by the high-level MPC and is not part of the decision variables.

A similar approach to ours has been adopted for controlling robotic systems with contact in \cite{kulchenko2011first}, where they showed successful ball-bouncing task. Considering the switching manifold to be the end of the horizon, they formulated an MPC problem (named first-exit) up to some set of terminal states, and applied a final cost at the terminal states equal to the differential cost-to-go for the infinite-horizon problem. However, they proposed some domain-specific heuristics to find the terminal cost. Our approach has fundamental differences with respect to \cite{kulchenko2011first}, i.e. 1) We propose a set of viability-based terminal cost and constraint, where the effects of constraints after the terminal state are also taken into account 2) We use a two-stage constrained MPC problem with viability guarantees, compared to \cite{kulchenko2011first} that used unconstrained iLQG 3) we use a highly-underactuated biped robot, and demonstrate an extensive set of real-world experiments with various disturbances.

The main contributions of the paper are as follows:
\begin{itemize}
    \item We propose a two-level VH-MPC framework that takes into account both the CoM and swing foot dynamics to control bipedal walking.
    \item We compare our proposed swing foot trajectory generation to polynomial based approach used in the literature.
    \item We demonstrate an MPC walking controller with both step location and timing adaptation on a real biped robot with passive ankles.
    \item We present walking experiments on uneven, soft, and slippery surfaces using our two-level VH-MPC on a robot without actuated ankles.
\end{itemize}

\begin{figure}[ht]
    \centering
    \includegraphics[width=1.\linewidth]{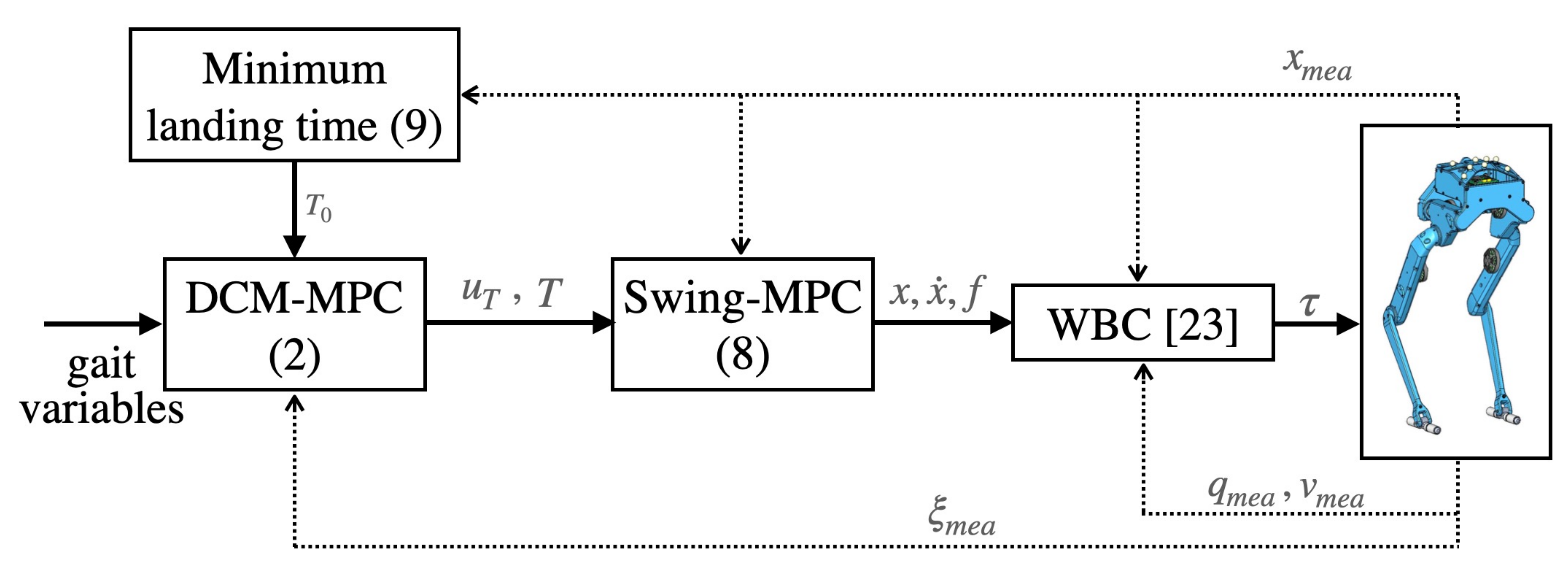}
    \caption{Block diagram of the control pipeline}
    \label{fig:block_diagram}
\end{figure}

A block diagram of the full control pipeline proposed in this paper is shown in Fig. \ref{fig:block_diagram}. The paper is structured as follows:
Section \ref{sec:high_level_MPC} briefly summarizes the high-level MPC problem that finds optimal step location and timing. In Section \ref{sec:swing_MPC}, we present the procedure of projecting the robot dynamics to the swing foot and formulating its corresponding MPC problem. In Section \ref{sec:framework_summary}, we summarize the whole control pipeline we use in the paper. In Section \ref{sec:results}, we present an extensive set of simulation and experimental results. Finally, Section \ref{sec:conclusions} concludes the findings.

\section{Foot location and timing adjustment}\label{sec:high_level_MPC}

Using the CoM and DCM as states, the LIPM dynamics can be written as \cite{englsberger2015three}
\begin{subequations}
\label{eq:DCM_CoM}
\begin{align}
&\dot{\com} = \omega_0 (\dcm-\com) \label{eq:DCM_CoM_1} \\ 
&\dot{\dcm} = \omega_0 (\dcm-\cop) \label{eq:DCM_CoM_2}
\end{align}
\end{subequations}
where $\com \in \mathbb{R}^2$ is the CoM horizontal position (CoM height has a fixed value $z_0$), and $\dcm=(\com+\dot{\com}/\omega_0)  \in \mathbb{R}^2$ is the 2D-DCM. $\cop\in\mathbb{R}^2$ is the CoP position, and in case of point contact feet, $\cop$ is identical to the contact point. $\omega_0$ is the natural frequency of the pendulum ($\omega_0=\sqrt[]{g/z_0}$, where $g$ is the gravity constant).
Equation \eqref{eq:DCM_CoM} explicitly separates the stable and unstable parts of the LIPM dynamics, where the CoM converges to the DCM \eqref{eq:DCM_CoM_1} and the DCM is pushed away by the CoP \eqref{eq:DCM_CoM_2}. 

To stabilize the DCM dynamics, we proposed in \cite{khadiv2020walking} to adapt the next step location and timing of the swing foot based on the DCM measurement $\dcm_{mea}$ using
\begin{subequations}
\label{eq:DCM_MPC}
\begin{align}
\underset{\nextstep, \Gamma, \offset}{\text{minimize}} \quad  & \alpha_1 \Vert \nextstep-\cop- \begin{bmatrix} l_{nom} \label{eq:MPC_cost}\\
w_{nom}\end{bmatrix}  \Vert^2  + \alpha_2 |\Gamma-\Gamma_{nom}|^2 \nonumber\\
& + \; \alpha_3  \Vert \offset-\begin{bmatrix} b_{x,nom}\\b_{y,nom}\end{bmatrix} \Vert^2\\ \nonumber \\
\text{s.t.} \qquad &\begin{bmatrix} l_{min}\\w_{min} \end{bmatrix}\leq \nextstep-\cop \leq  \begin{bmatrix} l_{max}\\
w_{max} \end{bmatrix} \label{eq:MPC_constraint_location}\\
&e^{\omega_0 T_{min}}\leq \Gamma \leq  e^{\omega_0 T_{max}}\label{eq:MPC_constraint_time}\\
&\nextstep+\offset=(\dcm_{mea}-\cop) e^{-\omega_0 t}\Gamma+\cop \label{eq:MPC_constraint_dynamics}\\
& \Gamma \geq \Gamma_0 \: e^{\omega_0 t} \label{eq:MPC_constraint_min_time}\\
&\begin{bmatrix} b_{x,min}\\b_{y,max,out} \end{bmatrix}\leq \offset \leq \begin{bmatrix} b_{x,max}\\b_{y,max,in} \end{bmatrix} \label{eq:MPC_constraint_viability}
\end{align}
\end{subequations}

where $\offset=\dcm_T - \nextstep$ is the DCM offset, $\Gamma = e^{\omega_0 T}$ is an exponential transformation of the step time $T$. $\cop$ and $\nextstep$ are the current and next step locations, while $l$ and $w$ are the step length and width, respectively. $\Gamma_0 = e^{\omega_0 T_0}$, where $T_0$ is a minimum time required for the swing foot to touch the ground, given its current state (see section \ref{subsec:min_time} for computing $\Gamma_0$). Subscript $nom$ stands for the nominal value and $in$ and $out$ are the inward and outward lateral directions \cite{khadiv2020walking}.

The three cost terms in \eqref{eq:MPC_cost} try to bring the gait variables to their nominal values which correspond to a desired walking velocity. Constraints \eqref{eq:MPC_constraint_location} , \eqref{eq:MPC_constraint_time} are box constraints on the location and (exponential of) time of the step. Equation \eqref{eq:MPC_constraint_dynamics} is the DCM dynamics which is linear as a function of next step location and (exponential of) time and DCM offset.  \eqref{eq:MPC_constraint_min_time} is a new constraint we define in this paper to prevent the planner to ask for an instantaneous step in the presence of uncertainties. 
Finally, \eqref{eq:MPC_constraint_viability} is the viability constraint on the DCM offset which is implemented in practice as a soft constraint to make sure that the program always find a feasible solution, even if the system is not viable \cite{khadiv2020walking}. Note that since we consider \textit{contact switch} as the terminal state of \eqref{eq:DCM_MPC} (and not a fixed horizon starting from the current time), the tail of the optimal control problem can be written using \textit{only the boundary of the gait values}, i.e. $T_{min}$ and $l_{max}$ to find viability bounds $b_{min/max}$. Hence, we can write state-independent viability constraint at the end of the horizon which ensures viability (weak forward invariance) of the gait.

The last cost term and constraint on the DCM offset have crucial role in our walking controller. The former incentivizes the DCM offset toward a desired offset that corresponds to a desired walking velocity, while the latter ensures viability of the gait, i.e. there exist at least a set of feasible steps into the future by which the robot can prevent a fall. Solving \eqref{eq:DCM_MPC} at each control cycle using the current measurement of the DCM yields the next step location and timing that should be realized by the swing foot trajectory. Since the step time is a decision variable, the horizon can be potentially adapted to preserve viability. Hence, \eqref{eq:DCM_MPC} as the high-level MPC problem of our framework is a VH-MPC.

\section{Projected swing foot dynamics}\label{sec:swing_MPC}
In this section, we examine the structure of the robot dynamics projected to the swing foot space. The main goal is to find a dynamic model with minimum simplification based on which we adapt the swing foot trajectories in real-time. 

The dynamics of a floating-base system can be written as 
\begin{equation}\label{eq:rigid_body}
    M(\gencoor) \dot \genvel + \nonlinear(\gencoor,\genvel) = B \torque + J_c^T \force
\end{equation}
where $M \in \mathbb{R}^{(n+6) \times (n+6)}$ is the robot mass matrix, $\gencoor \in \mathit{SE}(3) \times \mathbb{R}^n$ denotes the configuration space, $\genvel \in \mathbb{R}^{n+6}$ encodes the vector of generalized velocities (or more precisely quasi-velocities \cite{baruh1999analytical}), $\nonlinear \in \mathbb{R}^{n+6}$ is a concatenation of nonlinear terms including centrifugal, Coriolis and gravitational effects. $B \in \mathbb{R}^{(n+6) \times n}$ is a selection matrix that separates the actuated and unactuated Degrees of Freedom (DoFs), $\torque \in \mathbb{R}^n$ is the vector of actuating torques, $J_c \in \mathbb{R}^{3m \times (n+6)}$ is the Jacobian of $m$ foot in contact, and finally $\force \in \mathbb{R}^{3m}$ is the vector of contact forces (here we assume point-contact feet).

For bipedal walking, we assume one of the robot's feet is in stationary contact with the ground. With this assumption, we can write down the constraint-consistent projected dynamics of the robot to the swing foot as \cite{mistry2012operational}
\begin{equation}\label{eq:swing_foot}
    \underbrace{\Lambda_c \ddot \foot}_\text{Inertia} + \underbrace{\Lambda_c(J M_c^{-1} P \nonlinear - (\dot J + J M_c^{-1} \dot P) \genvel)}_\text{nonlinear terms} = \footforce
\end{equation}

where $\foot \in \mathbb{R}^{3}$ is the swing foot position and $P$ is the orthogonal projection operator $I - J_c^{\dagger} J_c$ (where $^{\dagger}$ stands for the Moore-Penrose inverse) which is a mapping to the nullspace of the contact constraint and $\dot P \genvel=- J_c^{\dagger} \dot J_c \genvel$. Note that $J \in \mathbb{R}^{3 \times (n+6)}$ is the swing foot Jacobian which is different from $J_c$ (the stance foot Jacobian). The constraint inertia matrix is denoted by $M_c=PM + I - P$ and $\Lambda_c=(J M_c^{-1} P J^T)^{-1}$ is the apparent mass at the swing foot. In this equation, $\footforce$ is the the \textit{swing foot actuation force}, i.e. if the swing foot were to apply a force (virtually) to the environment, how much force it could exert without violating the stance foot constraints and it can be derived as
\begin{equation}\label{eq:swing_force}
\footforce = \Lambda_c J M_c^{-1} P B \torque
\end{equation}
Given the friction cone constraints of the stance foot and joint torque limits, the actuation force to move the swing foot is constrained. In the sequel we find an approximation of these constraints and a simplified dynamics of the swing foot that can be used to generate dynamically consistent swing foot trajectories.

\subsection{Simplified swing foot dynamics}\label{subsec:simplified_swing_dynamics}
The swing foot dynamics \eqref{eq:swing_foot} is highly nonlinear as a function of the robot configuration. As a result, using this dynamic model in MPC leads to a nonlinear and non-convex optimization problem that needs several iterations to be solved and it might get stuck in undesired local minima. However, we need to be able to regenerate the swing foot trajectory quickly based on the updated landing location and time from the high-level DCM planner. Hence, we resort to a simplified linear dynamics of the foot in the reachable area of the swing foot. 

To be able to compare the effects of different terms in \eqref{eq:swing_foot}, we used the approach in \cite{khadiv2020walking} and generated different swing foot trajectories using polynomials in simulation for different walking velocities. Note that we did all the simulations on the robot we study in this paper, Bolt (the robot weight is roughly \SI{12.5}{N}). We exerted to the robot CoM random disturbances at the start of each step, $\SI{-2}{N} < F_x,F_y< \SI{2}{N}$ and $\SI{-1}{N} < F_z < \SI{1}{N}$, and reset the simulation if the robot would fall down. We used the whole body controller (WBC) in \cite{grimminger2020open} to map the trajectories to joint torques and apply them in simulation. With this strategy, We generated 1300 samples (corresponding to 50 walking steps with different timings) to approximate the swing foot dynamics. For these experiments, we set the following bounds for steps : $l_{min} = \SI{-0.12}{m}$, $l_{max} = \SI{0.12}{m}$, $w_{min} = \SI{-0.1}{m}$, $w_{max} = \SI{0.3}{m}$, $T_{min} = \SI{0.1}{s}$, $T_{max} = \SI{0.3}{s}$. Furthermore, we randomly changed the nominal step time in the range $T_{min} < T_{nom} < T_{max}$. Due to the exerted random disturbances, the robot needs to step in different directions while respecting the kinematic constraints.

\begin{figure}[ht]
\centering
  \includegraphics[width=1.\linewidth]{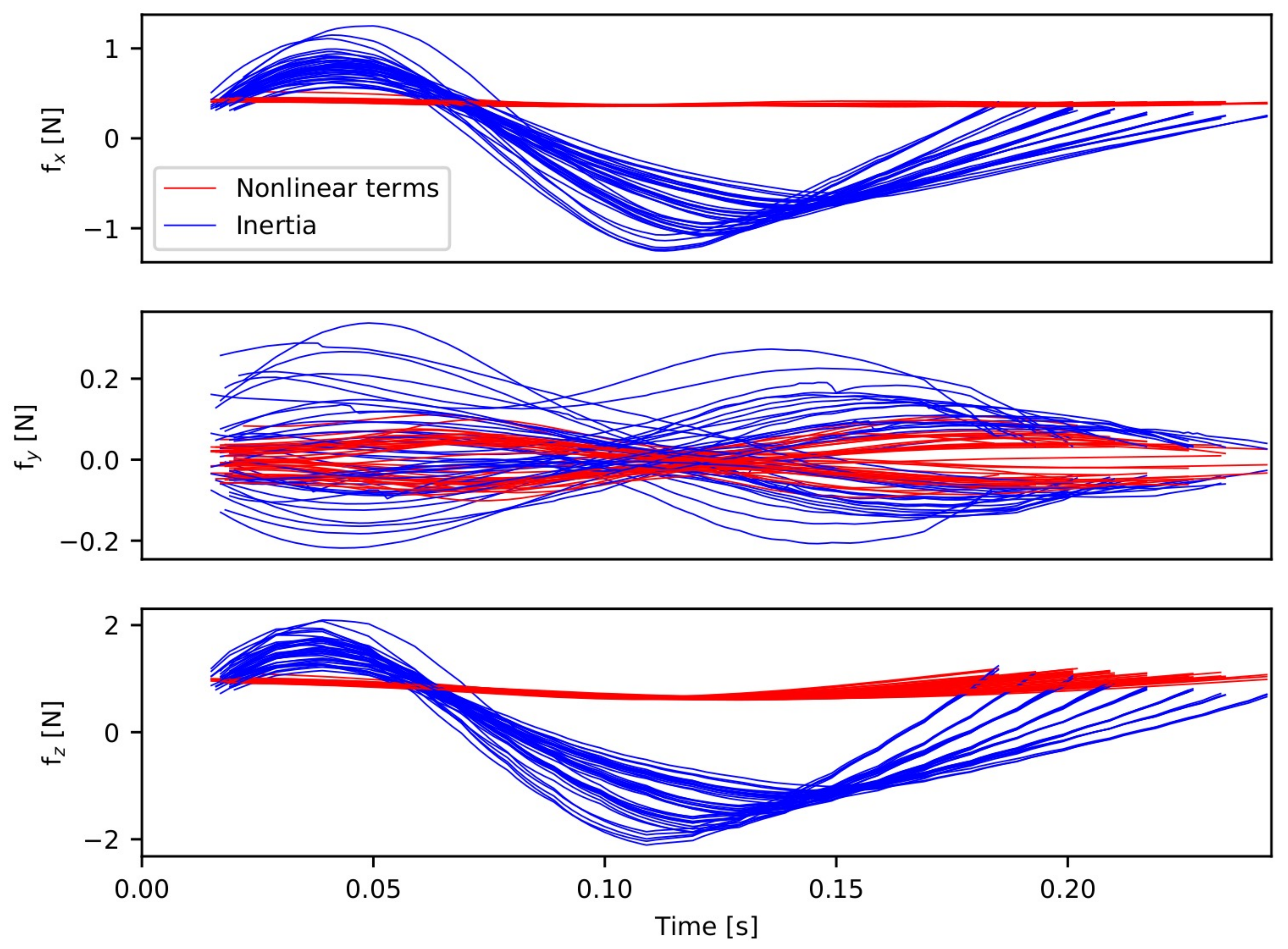}
  \caption{inertial vs nonlinear effects at the swing foot for a set of 50 different landing location and time}
  \label{fig:nonlinear_vs_inertia}
  \vspace{-0.5cm}
\end{figure}

Using the state trajectories at each time, we compute the effects of different terms. 
In Fig. \ref{fig:nonlinear_vs_inertia} we show the contribution of the inertia and nonlinear terms to the swing foot dynamics for 50 different swing phase (different landing location and time). 
As we can see in Fig. \ref{fig:nonlinear_vs_inertia}, inertia is the main effect in swing foot dynamics while the nonlinear terms remain relatively constant. In an effort to simplify the swing foot dynamics, we approximate all the nonlinear terms with a constant term $\nonlinearconstant$.

\begin{figure}[ht]
\centering
  \includegraphics[width=1.\linewidth]{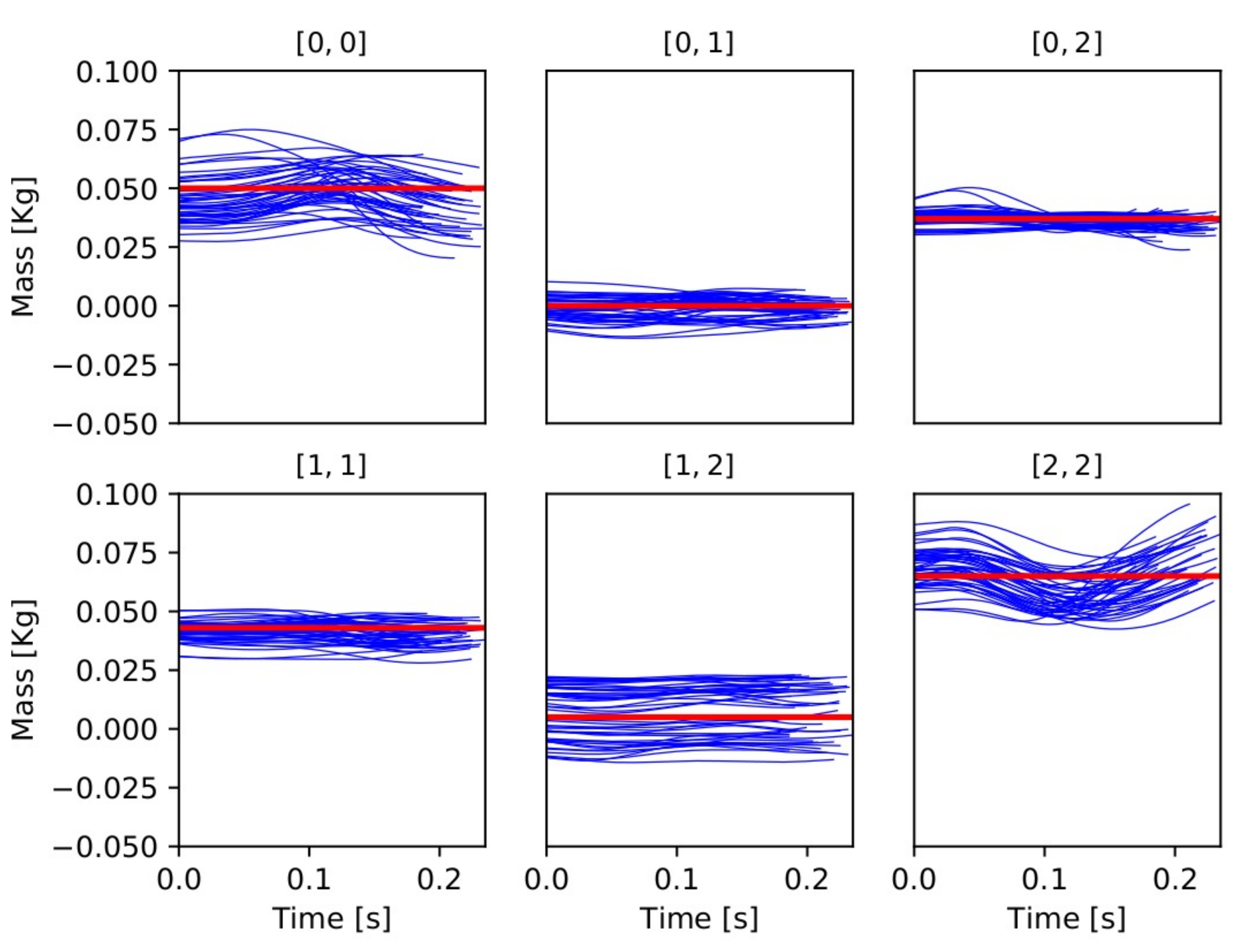}
  \caption{Mass matrix elements for 50 different landing location and time; the mean value of each component is specified by red.}
  \label{fig:mass_matrix}
  \vspace{-0.5cm}
\end{figure}

Now, we inspect further the swing foot mass matrix structure and plot its variation as a function of the robot configuration, in the range of swing foot motion that we are interested in. As the mass matrix is symmetric, we only need to examine the diagonal and upper(or lower)-triangular part of the matrix. In Fig. \ref{fig:mass_matrix}, we plotted the components of the swing foot mass matrix, for the same set of motions we used to compare inertial and nonlinear effects. As we can see in this figure, the projected mass matrix components do not change significantly during one step, hence we propose to use the following linear dynamics to be used inside a linear MPC problem for the swing foot.

\begin{equation}\label{eq:swing_foot_dynamics}
    \footforce = \Lambda \ddot \foot + \nonlinearconstant
\end{equation}

where the components of $\Lambda$ and $\nonlinearconstant$ are constant. 

\subsection{Constraints on the swing foot actuation force}\label{subsec:swing_constraints}
In this section, we use the intermediate variable swing foot actuation force $\footforce$ to denote the feasible actuation that can be applied to the swing foot, given the joint torque limits and the stance foot friction cone constraints. By defining this variable, we would like to emphasise that the mass matrix projected to the swing foot is not diagonal in general (as we have seen in Section \ref{subsec:simplified_swing_dynamics}) and enforcing constraints on the swing foot acceleration in different directions does not account for this. In fact, a linear combination of the swing foot accelerations should be limited, where this coupling is given by the swing foot mass matrix.

The idea is to use the set of different configurations we considered in the previous section and construct the following Linear program (LP) to find the boundaries of the forces that can be applied by the swing foot without violating any of the constraints at each configuration
\begin{subequations}
	\label{eq:swing_force_limit}
	\begin{align}
	    \label{eq:cost_max_force}
	   \underset{\torque}{\text{minimize}} \quad &cost   \triangleq \selection^{T} \footforce = \selection^{T} \Lambda_c J M_c^{-1} P B \torque\\
	   \label{eq:unilaterality}
	    \text{s.t.} \quad&\rho_z + \eta_z \ge 0 \\
	    \label{eq:friction_cone_x}
	    \qquad & |(\rho_x+\eta_x)| \leq \frac{\sqrt{2}}{2} \mu (\rho_z + \eta_z) \\
	    \label{eq:friction_cone_y}
	    \qquad & |(\rho_y+\eta_y)| \leq \frac{\sqrt{2}}{2} \mu (\rho_z + \eta_z) \\
	    \label{eq:torque_limit}
	    \qquad &\torque_{min} \le \torque \le \torque_{max}
	\end{align}
\end{subequations}

In this equation, $\rho = -(J_c^T)^{\dagger}(I-P)((I-M M_c^{-1}P) \nonlinear+M M_c^{-1} \dot P \genvel)$ and $\eta=-(J_c^T)^{\dagger}(I-P)(I-M M_c^{-1}P)B\torque$ \cite{xin2020optimization}. $\mu$ is the friction coefficient and
$\selection \in \mathbb{R}^{3}$ in \eqref{eq:cost_max_force} is a selection vector whose one of elements is either $+1$ or $-1$, and the rest are zero. By iterating over all six possibilities, we can compute the maximum and minimum actuation force at the swing foot in each direction, for a given configuration. The constraints of the problem are unilaterality of contact \eqref{eq:unilaterality}, linearized friction cone constraints of the stance foot \eqref{eq:friction_cone_x}, \eqref{eq:friction_cone_y}, and the joint torque limits \eqref{eq:torque_limit}.  

\begin{figure}[ht]
\centering
  \includegraphics[width=1.\linewidth]{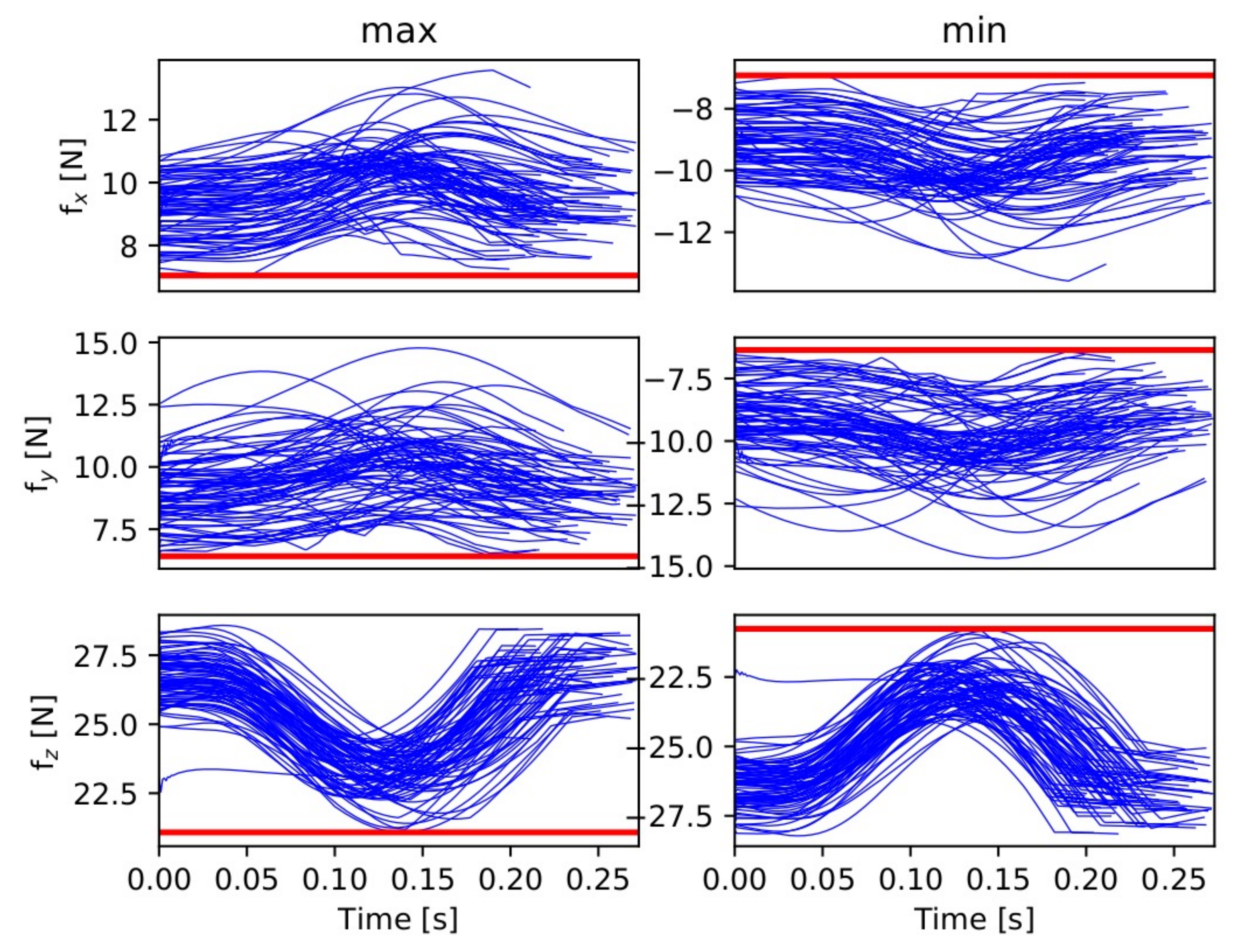}
  \caption{Maximum and minimum actuation force at the swing foot for 50 different landing location and time; the worst case of these approximated upper/lower bounds is specified by red.}
  \label{fig:min_max}
\end{figure}

We solved \eqref{eq:swing_force_limit} for the same set of trajectories we used in Section \ref{subsec:simplified_swing_dynamics}. Figure \ref{fig:min_max} shows the time history of the maximum and minimum of the swing foot actuation force in each direction. We also found the worst-case of the approximate bounds in \eqref{eq:swing_force_limit}  over all explored configurations and specified by red lines in Fig. \ref{fig:min_max}. The computed state-independent bounds on the maximum and minimum forces will be used inside an MPC problem in the next subsection to generate swing foot trajectories. Note that for solving \eqref{eq:swing_force_limit} we set the friction coefficient to $\mu=0.5$, and the minimum and maximum joint torques to $\SI{-2}{\momentunit}$ and $\SI{2}{\momentunit}$ which are robot-dependent. 

\begin{remark}
What we did in Section \ref{subsec:simplified_swing_dynamics} and \ref{subsec:swing_constraints} is similar to \cite{bledt2020extracting} in that we simulate the robot using a candidate controller and use the generated data to find interpretable heuristics. However, as opposed to \cite{bledt2020extracting} that extracts heuristics to regularize the optimization problem cost function for different desired behaviors, we find a linear model in the swing foot space (with corresponding constraints) and use it to find dynamically consistent swing foot trajectories. Furthermore, in this paper we are tackling the bipedal locomotion problem with passive ankles which is far more unstable than a quadruped locomotion problem. Hence, it is crucial to control the swing foot landing location and time \emph{precisely} to have a stable gait.
\end{remark}

\subsection{Proposed foot trajectory generator}\label{sec:low_level_MPC}


In this section, we formulate the swing foot reactive trajectory generation problem as a VH-MPC. The main goal of this MPC is to bring the swing foot on the ground at the desired time computed by \eqref{eq:DCM_MPC} and to be as close as possible to the desired foot location.


We use the simplified swing foot dynamics in \eqref{eq:swing_foot_dynamics} and the approximation of the force boundaries computed based on \eqref{eq:swing_force_limit}.
We write then the following VH-MPC to find optimal force applied to the swing foot to land on the ground at a desired time and as close as possible to the desired location
\begin{subequations}
	\label{eq:swing_MPC}
	\begin{align}
	    \label{eq:swing_cost}
	    \underset{\footforce_i}{\text{minimize}} \:\: &\sum\limits_{i=1}^{N}\, \alpha_1 \Vert \footforce_i \Vert^2  + \alpha_2  \Vert \foot_N-\foot_{f} \Vert^2+ ... \nonumber\\
	    &\qquad \alpha_3  \Vert \dot \foot_N- \dot \foot_{f} \Vert^2 + \alpha_4  \Vert z(\frac{T}{2})- z_{mid} \Vert^2\\
	    \label{eq:swing_x_dot}
	    \text{s.t.} \quad \foot_i &=A^i \foot_{mea}+ \sum\limits_{j=1}^{i}\, 
                                A^{i-j} B \big((\footforce_j - \nonlinearconstant) \Lambda_c^{-1} \big)^T\\
	    \label{eq:constraint_swing_force}
	    \qquad &\footforce_{min} \le \footforce_{i} \le \footforce_{max} \quad , \quad \forall i=1,...,N. \\
	    \label{eq:swing_height}
	    \qquad &z_{min} \le z_{i} \le z_{max}  \quad , \quad \forall i=1,...,N.\\
	    \label{eq:swing_terminal_height}
	    \qquad &z_f =0 \: , \: \dot z_f =0	    
	\end{align}
\end{subequations}

where $A$ and $B$ are obtained, writing \eqref{eq:swing_foot_dynamics} in standard discrete state space form.
$\footforce_i$ is the applied force at the swing foot at $i$'th node in the horizon, $\foot=[x,y,z]^T$ and $\dot \foot=[\dot x, \dot y, \dot z]^T$ are the swing foot position and velocity and subscripts  $f$ and $mea$ stand for the end of the current step (horizon) and the measured value, respectively. $z_{mid}$ is the desired step height. Note that in this program, only the terminal state of the swing foot in vertical direction is a constraint which ensures landing at the \emph{desired time}.

\subsection{The minimum landing time}\label{subsec:min_time}
Depending on the current state of the swing foot, a minimum time can be computed that ensures landing the foot on the ground, given the constraints on the swing foot actuation force. This minimum time is used in \eqref{eq:MPC_constraint_min_time} to make sure that \eqref{eq:DCM_MPC} does not ask for a contact time that is not possible to be realized by the swing foot in \eqref{eq:swing_MPC}. We compute this minimum time using the following program
\begin{subequations}
	\label{eq:minimum_time}
	\begin{align}
	    \label{eq:min_time_cost}
	    \underset{\footforce_i,i}{\text{minimize}} \quad & i\\
	    \label{eq:min_time_x_dot}
	    \text{s.t.} \quad \foot_i = A^i &\foot_{mea}+ \sum\limits_{j=1}^{i}\, 
                                A^{i-j} B \big((\footforce_j - \nonlinearconstant) \Lambda_c^{-1} \big)^T\\
	    \label{eq:min_time_force}
	    \qquad &\footforce_{min} \le \footforce_{i} \le \footforce_{max} \quad , \quad \forall i=1,...,N. \\
	    \label{eq:min_time_terminal_height}
	    \qquad &z_f =0 \: , \: \dot z_f =0	    
	\end{align}
\end{subequations}

Equation \eqref{eq:minimum_time} is a very simple mixed-integer program ($i$ is an integer variable) that can be solved very fast. In our experiments, we used a simple approach to solve this program; we start from zero and increase $i$ and solve \eqref{eq:minimum_time} until the program becomes feasible. In our experiments, such a brute-force approach to solve \eqref{eq:minimum_time} took in worst case around $\SI{1}{\mu s}$.



\section{Full control pipeline}\label{sec:framework_summary}
In this section, we summarize the full control pipeline we use in this paper. The main parts of our control pipeline are two VH-MPC problems \eqref{eq:DCM_MPC} and \eqref{eq:swing_MPC} that reactively update the next contact location and time, and accordingly the swing foot trajectory (see Fig. \ref{fig:block_diagram}). The generated trajectories then are fed into a WBC to compute the joint torques. We used in this paper the task-space impedance controller we proposed in \cite{grimminger2020open} as a WBC. Note that we use the coordinate frame attached to the stance foot as the reference frame and conduct all the computations in this frame at each step.


After specifying the desired walking velocity and nominal gait parameter (step length, width, time, ...), the first step of our framework uses \eqref{eq:minimum_time} to compute the minimum time required for the swing foot to land on the ground. Then, \eqref{eq:DCM_MPC} is executed to find the optimal step location and timing of the step, given the current measurement of the DCM. This stage is formulated as a VH-MPC where the MPC horizon (step timing) is a decision variable. The resulting step location and timing are then used to generate a swing foot trajectory using the swing foot MPC \eqref{eq:swing_MPC}. The goal in the swing foot MPC is to realize landing of the foot at the time that is planned by the first stage, as close as possible to the desired position. In this stage we also have a VH-MPC, but the horizon is not decided by the MPC itself and it is set by the first stage MPC. Finally, the resulting trajectories are passed to the WBC \cite{grimminger2020open} to compute joint torques (see Fig. \ref{fig:block_diagram}).

\begin{remark}
The main reason behind splitting the problem into two MPC problems is that one holistic MPC that finds both the swing foot trajectory and optimal step location and timing would result in either a mixed-integer program or a non-convex problem. A mixed-integer program is not suitable for real-time applications because of its combinatorial complexity. A non-convex optimization needs several iterations to converge and may get stuck in a poor local minimum. We prefer our two-level MPC despite the fact that it is sub-optimal, because 1) it can be solved in real-time inside MPC 2) it always finds a feasible solution using the minimum time \eqref{eq:minimum_time} as a constraint in \eqref{eq:DCM_MPC}. 
\end{remark}

\section{results}\label{sec:results}
In this section, we first present a comparison between our proposed approach for swing foot trajectory generation to a polynomial-based trajectory generation approach. Then, we show an extensive set of real robot experiments demonstrating the capability of our framework in generating robust walking motions in the presence of various disturbances and uncertainties. For all the experiments, we ran our proposed two-level MPC at $\SI{100}{Hz}$ and the WBC at $\SI{1}{kHz}$.

\subsection{Swing foot trajectories comparison}\label{subsec:result_swing}
\subsubsection{Case study}
To compare qualitatively the performance of our proposed MPC-based swing foot trajectory generation \eqref{eq:swing_MPC} to a standard polynomial-based one \cite{khadiv2020walking}, we consider a push recovery scenario in simulation. We did a simulation where the robot steps in place and is pushed laterally by an impulse of $\SI{10}{\impulseunit}$, at $t=\SI{0.6}{s}$. 

All components of the control pipeline are exactly the same, except the swing foot trajectory generation part. For the polynomial-based swing foot trajectory generation, we parameterize the trajectories in horizontal directions using minimum-jerk (5th order) polynomials and find the unique coefficients that connect the current position, velocity and acceleration of the swing foot to the end position with zero velocity and acceleration. For the vertical direction, we construct a QP with inequality constraints (the foot height should remain in a certain bound) to find the coefficients of the polynomial. In the case that the corresponding problem gets infeasible, we use the solution of the previous iteration and shift it by one sample time. Compared to the heuristics used in \cite{khadiv2020walking} that we would not update the gait values at the instances close to the end of the step, here we use this simple but more systematic and practical strategy of shifting the previous solutions and this was enough to always have a feasible solution in our experiments. 

\begin{figure}[ht]
\centering
  \includegraphics[width=1.\linewidth]{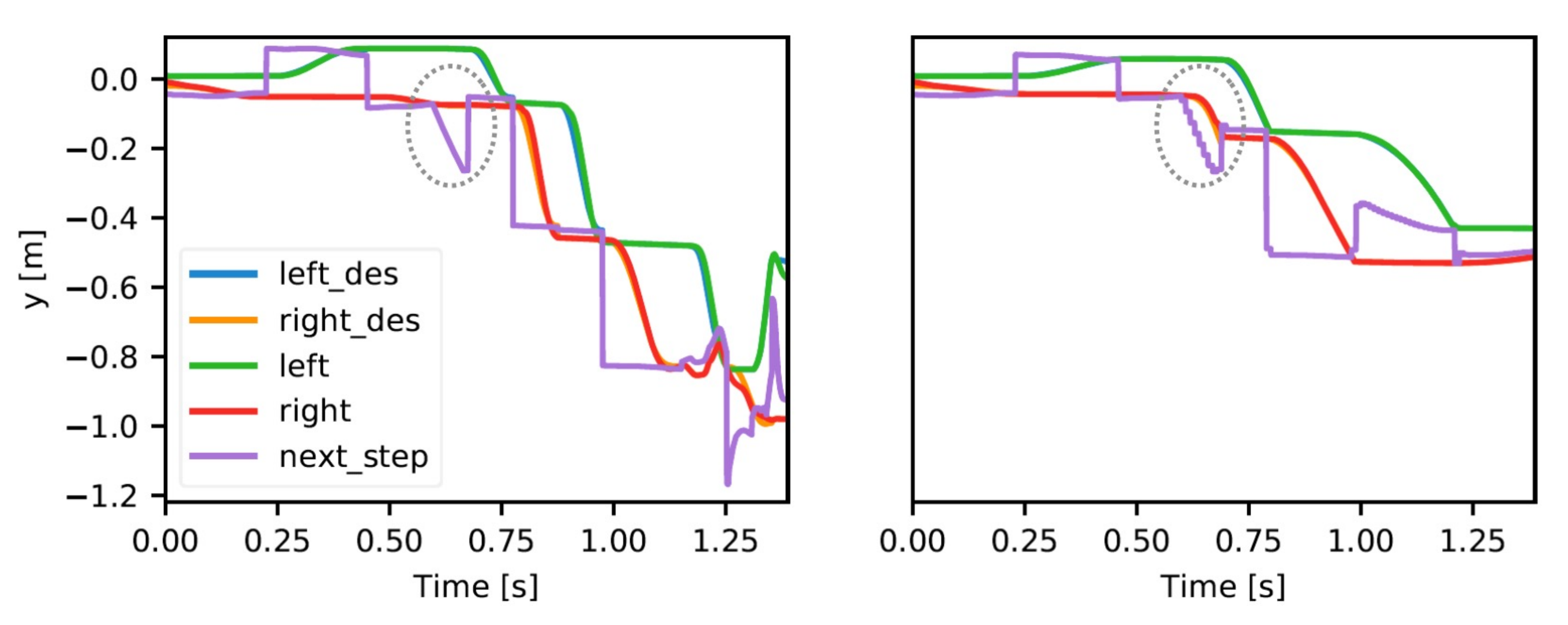}
  \caption{Comparison between swing foot trajectories in the $y$ direction generated using polynomials (left) and proposed MPC \eqref{eq:swing_MPC} (right).
}
  \label{fig:swing_horizontal}
\end{figure}

Figure \ref{fig:swing_horizontal} compares the resulting swing foot tracking performance of the two approaches. As we can see, after the push is exerted at $t=\SI{0.6}{s}$, the next step location is updated using \eqref{eq:DCM_MPC}. Since the polynomial-based swing foot trajectory generation (left figure) gets infeasible at this time, the solution in the previous iteration is used. This causes a discrepancy between the desire landing point asked by \eqref{eq:DCM_MPC} and the end point of the trajectory. This is enough for the robot to get unstable and fall down. However, using our novel MPC-based approach \eqref{eq:swing_MPC}, the swing foot is adapted after the disturbance as much as the constraints allow and the robot is able to recover from the disturbance.
%
%
\subsubsection{Systematic comparison}
To compare systematically the two approaches, we conducted 450 walking steps simulation with different pushes sampled inside the range of $\pm \SI{1.5}{\impulseunit}$ for the horizontal directions and $\pm \SI{0.75}{\impulseunit}$ in the vertical direction. 
Note that this range is small compared to the disturbance we exerted in the previous subsection ($\SI{10}{\impulseunit}$). In case of large disturbances, the robot might lose its balance and this would make it difficult to compare the landing location and time of the swing foot between two approaches. We already showed in the previous subsection that in the case of a large disturbance, our approach clearly outperforms the polynomial-based trajectory generation approach.

\begin{figure}[ht]
\centering
  \includegraphics[width=1.\linewidth]{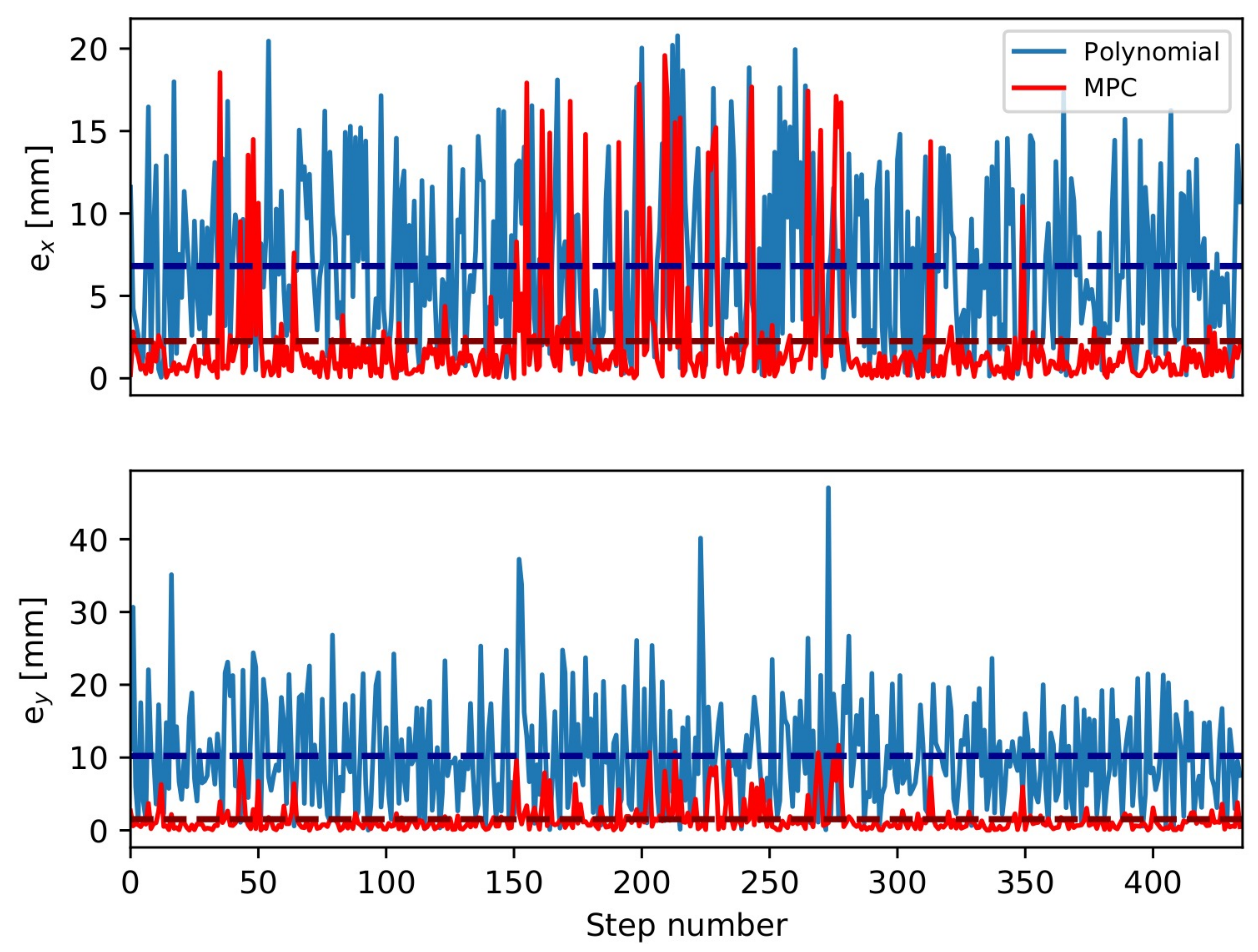}
  \caption{Landing location error for 450 walking steps simulation with random pushes sampled inside the range of $\pm \SI{1.5}{\impulseunit}$ for the horizontal directions and $\pm \SI{0.75}{\impulseunit}$ for the vertical direction. The mean values for each approach are shown by dashed lines with the same colour.}
  \label{fig:landing_error}
\end{figure}

\begin{figure}[ht]
\centering
  \includegraphics[width=1.\linewidth]{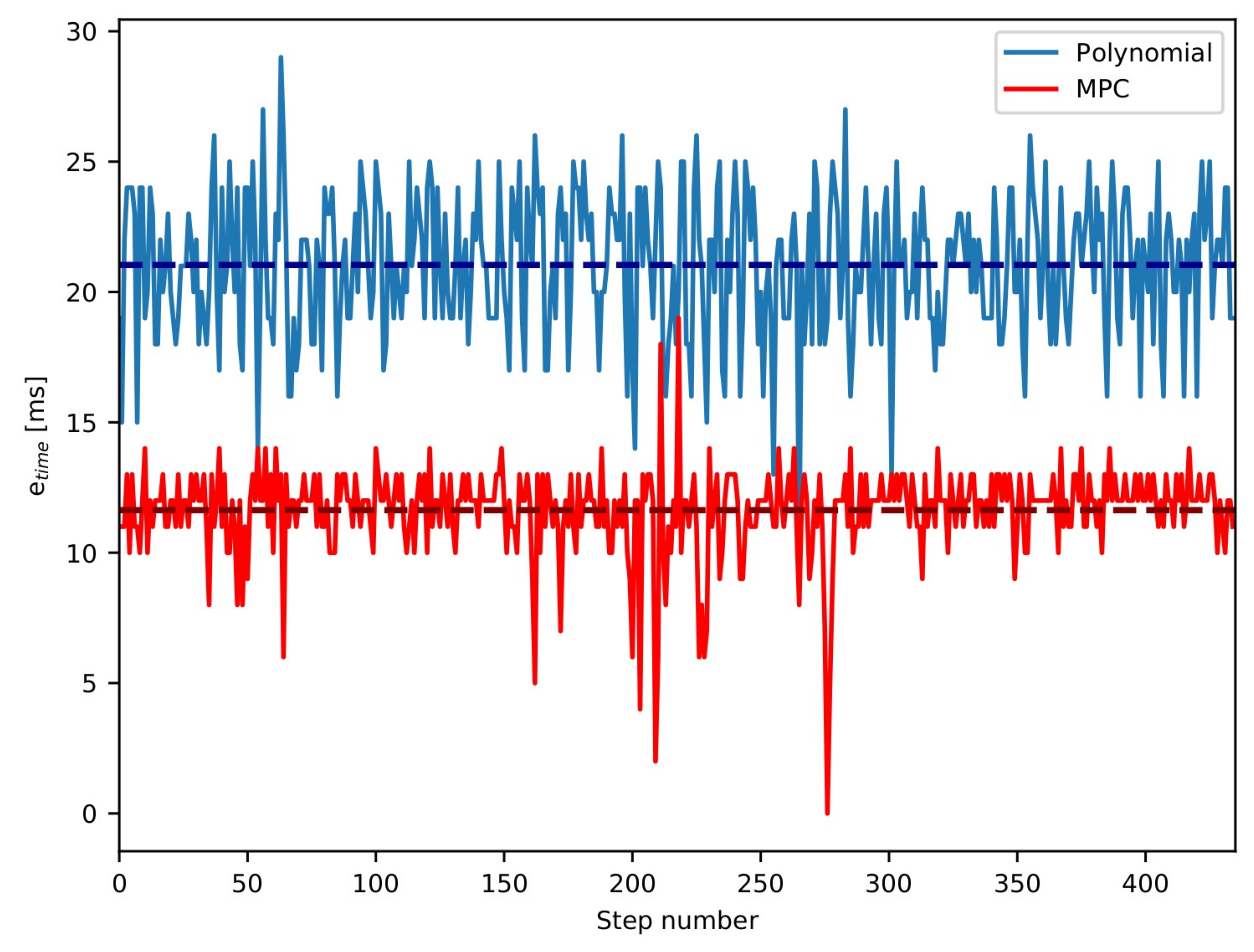}
  \caption{Landing time error for 450 walking steps simulation with random pushes sampled inside the range of $\pm \SI{1.5}{\impulseunit}$ for the horizontal directions and $\pm \SI{0.75}{\impulseunit}$ for the vertical direction. The mean values for each approach are shown by dashed lines with the same colour.}
  \label{fig:time_error}
\end{figure}

As we can see in Fig. \ref{fig:landing_error}, in both $x$ and $y$ directions, the error in landing location of our approach is less than the polynomial-based trajectory generation. Note that error in landing location can not only cause a degradation in velocity tracking (performance), but it may also jeopardize stability. More important than landing location error, landing time error of the swing foot in our approach is significantly lower than polynomial-based approach (see Fig. \ref{fig:time_error}). Landing time error has a very important role on realization of the gait, i.e. early landing can cause high impact forces, while late landing can cause high acceleration of the foot once the gait phase changes, given that the controller is not event-based.


\subsection{Real robot experiment}\label{subsec:result_experiment}
We implemented our proposed framework shown in Fig. \ref{fig:block_diagram} on Bolt, our new open-source biped robot. Bolt has been designed and fabricated using the 3D-printing technology and proprioceptive actuator concept that gives transparency in the drive system for torque-control proposed in \cite{grimminger2020open}. Each leg of Bolt has 3 active DoFs (2 in hip and one in knee) together with a passive ankle joint similar to \cite{kim2020dynamic}.

Bolt is capable of very dynamic movements and fast stepping, which renders the swing foot dynamics non negligible. We believe that this was the main reason we were not able to implement the polynomial-based swing foot trajectory generation on the real robot. In fact, as the statistical comparison showed in the previous subsection, the polynomial-based swing foot trajectory generation introduces large errors in the landing location and time of the swing foot which can explain our failure to implement it on the inherently highly unstable biped robot Bolt. Hence, in this subsection we only present and analyze different real robot demonstrations with the approach presented in the paper.

\begin{figure*}
	\centering
	\includegraphics[height=0.14\linewidth, trim={0cm 0cm 0cm 0cm}, clip]{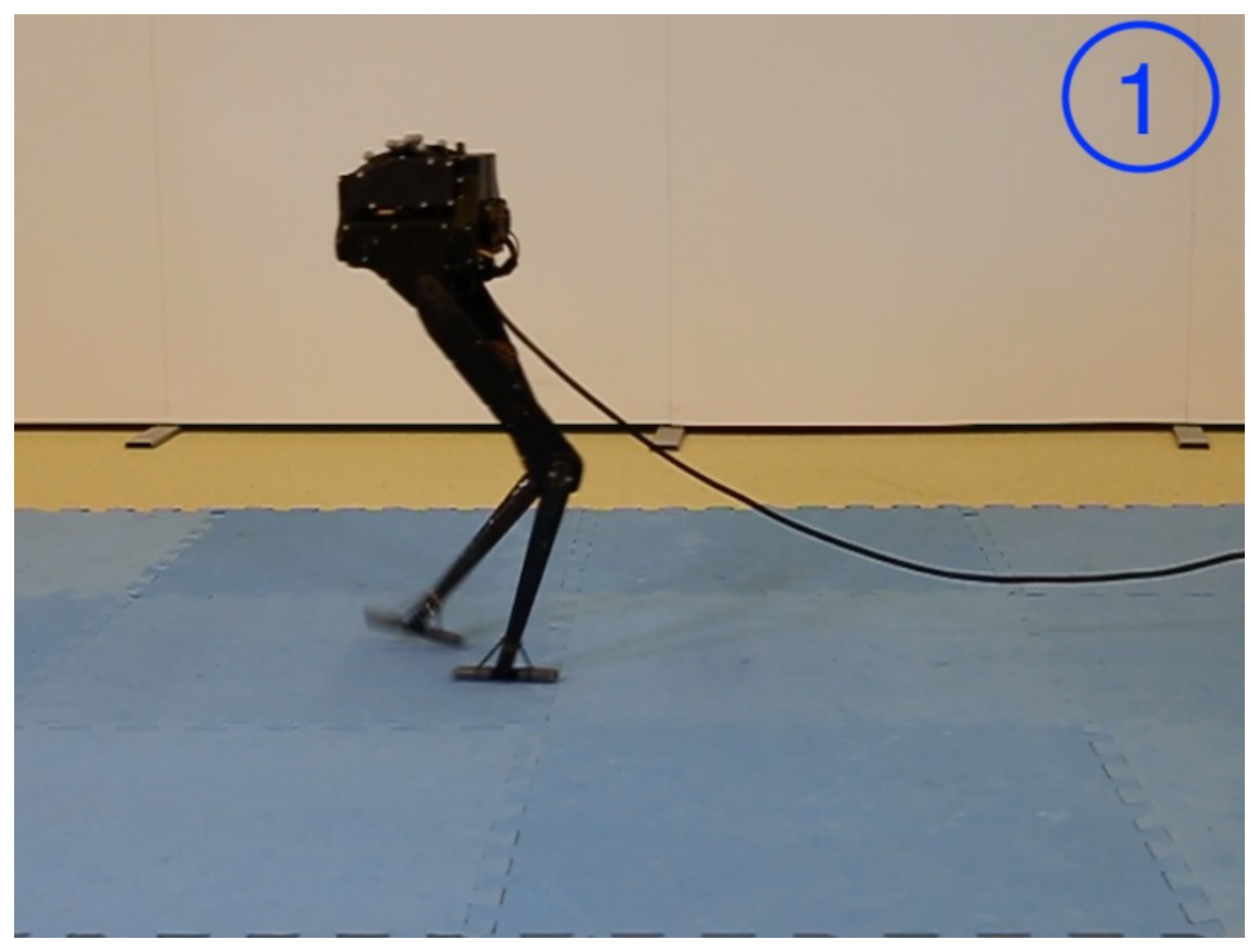}%
	\includegraphics[height=0.14\linewidth, trim={0cm 0cm 0cm 0cm}, clip]{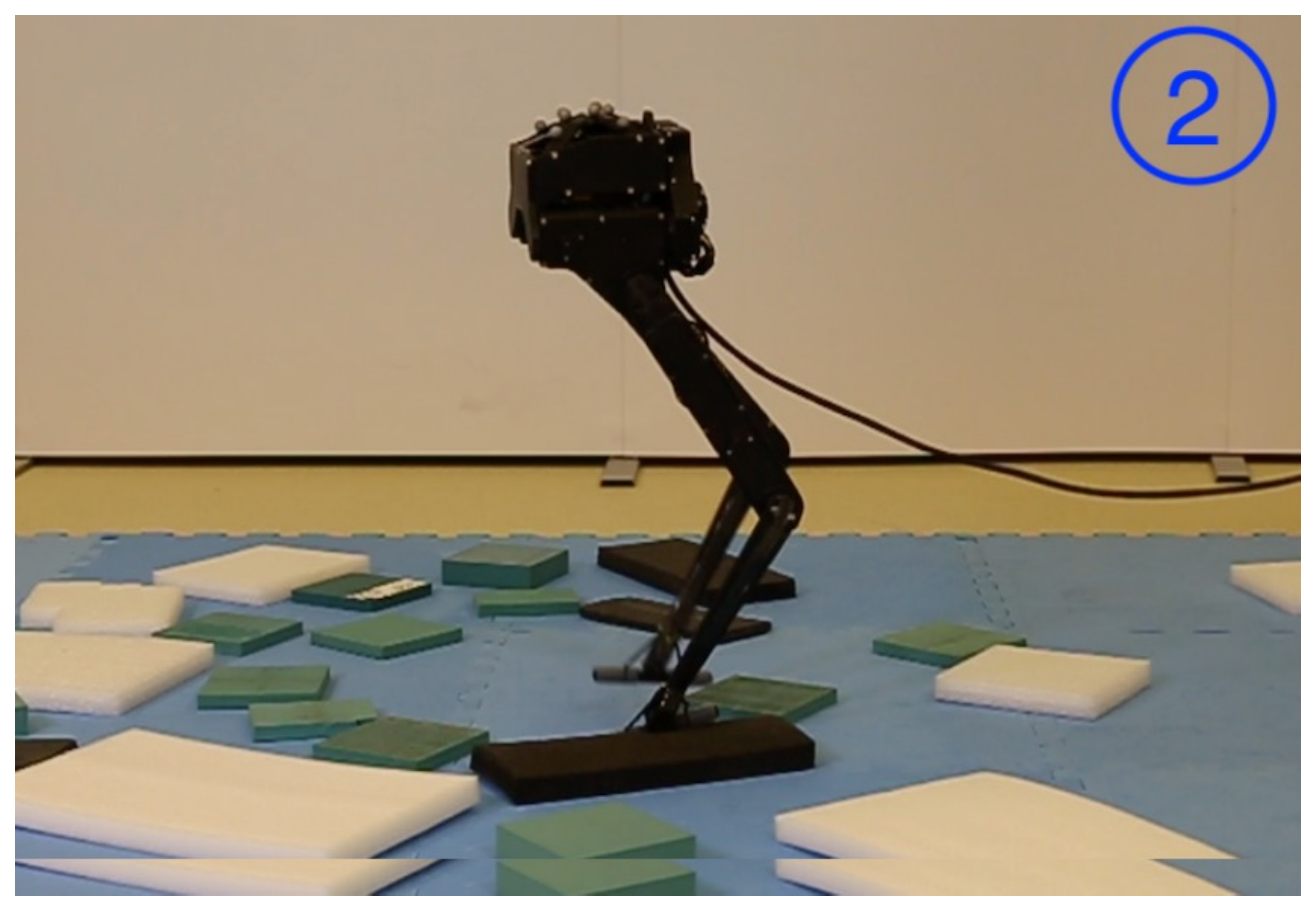}%
	\includegraphics[height=0.14\linewidth, trim={0cm 0cm 0cm 0cm}, clip]{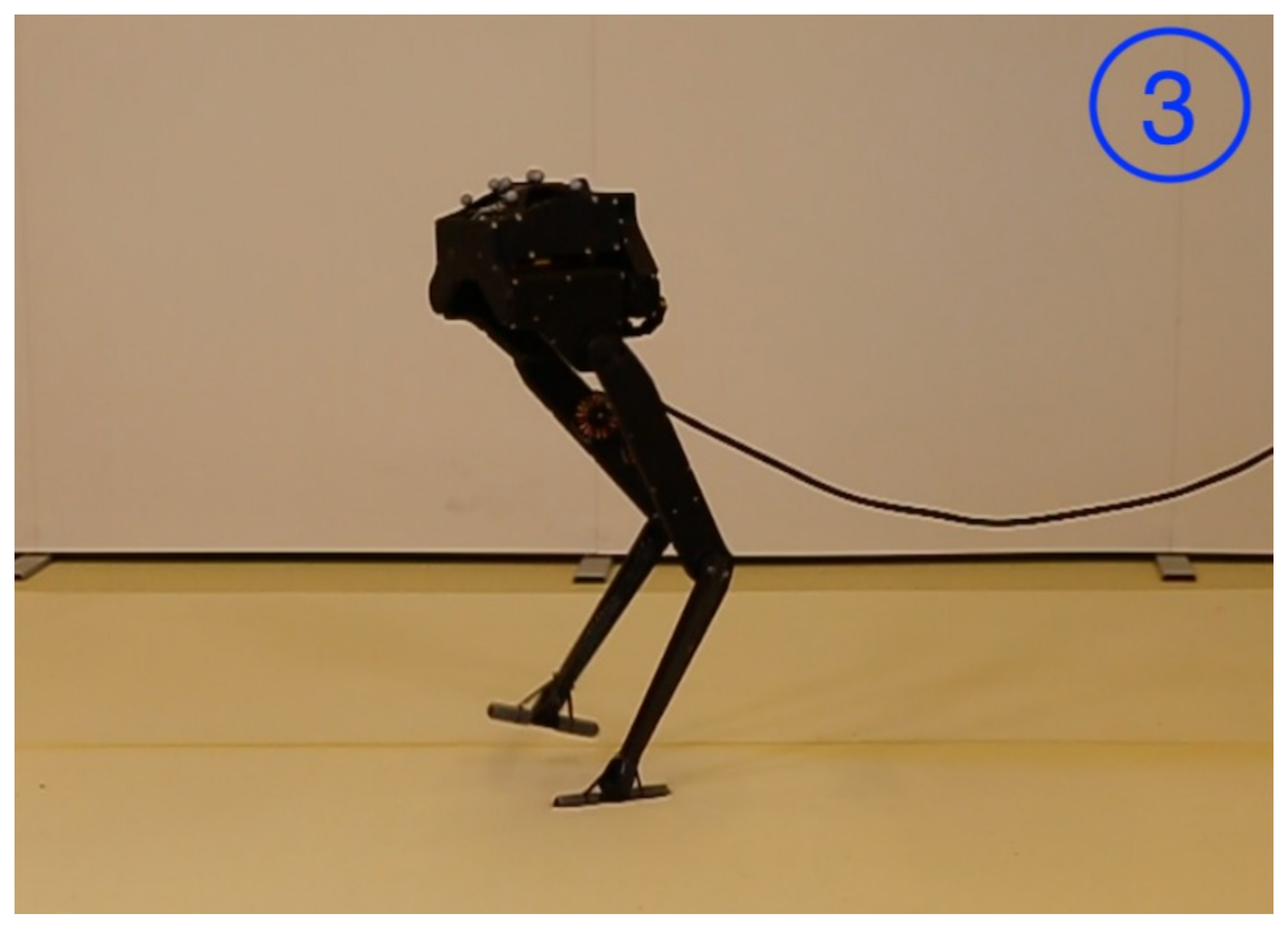}%
	\includegraphics[height=0.14\linewidth, trim={0cm 0cm 0cm 0cm}, clip]{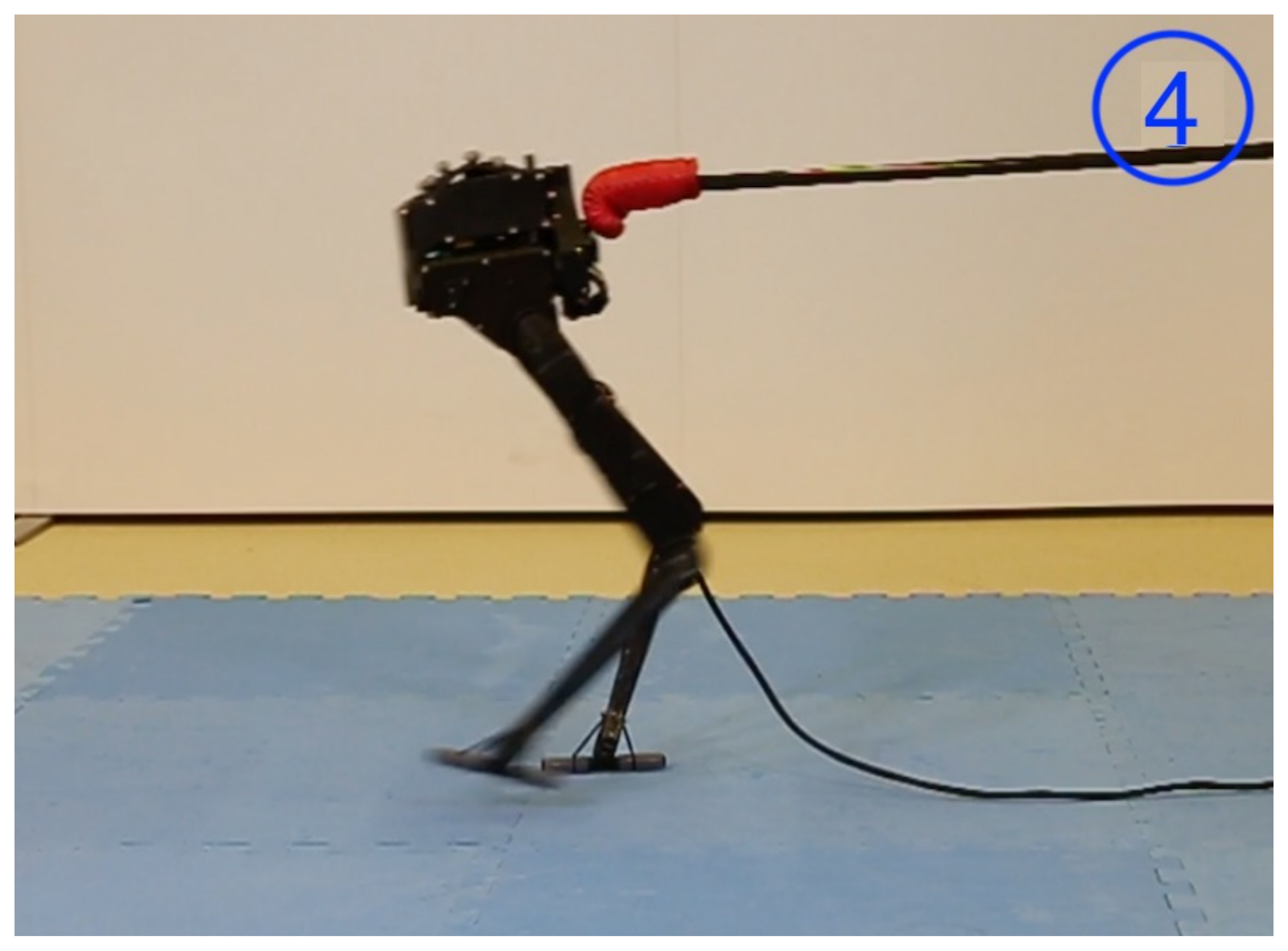}%
	\includegraphics[height=0.14\linewidth, trim={0cm 0cm 0cm 0cm}, clip]{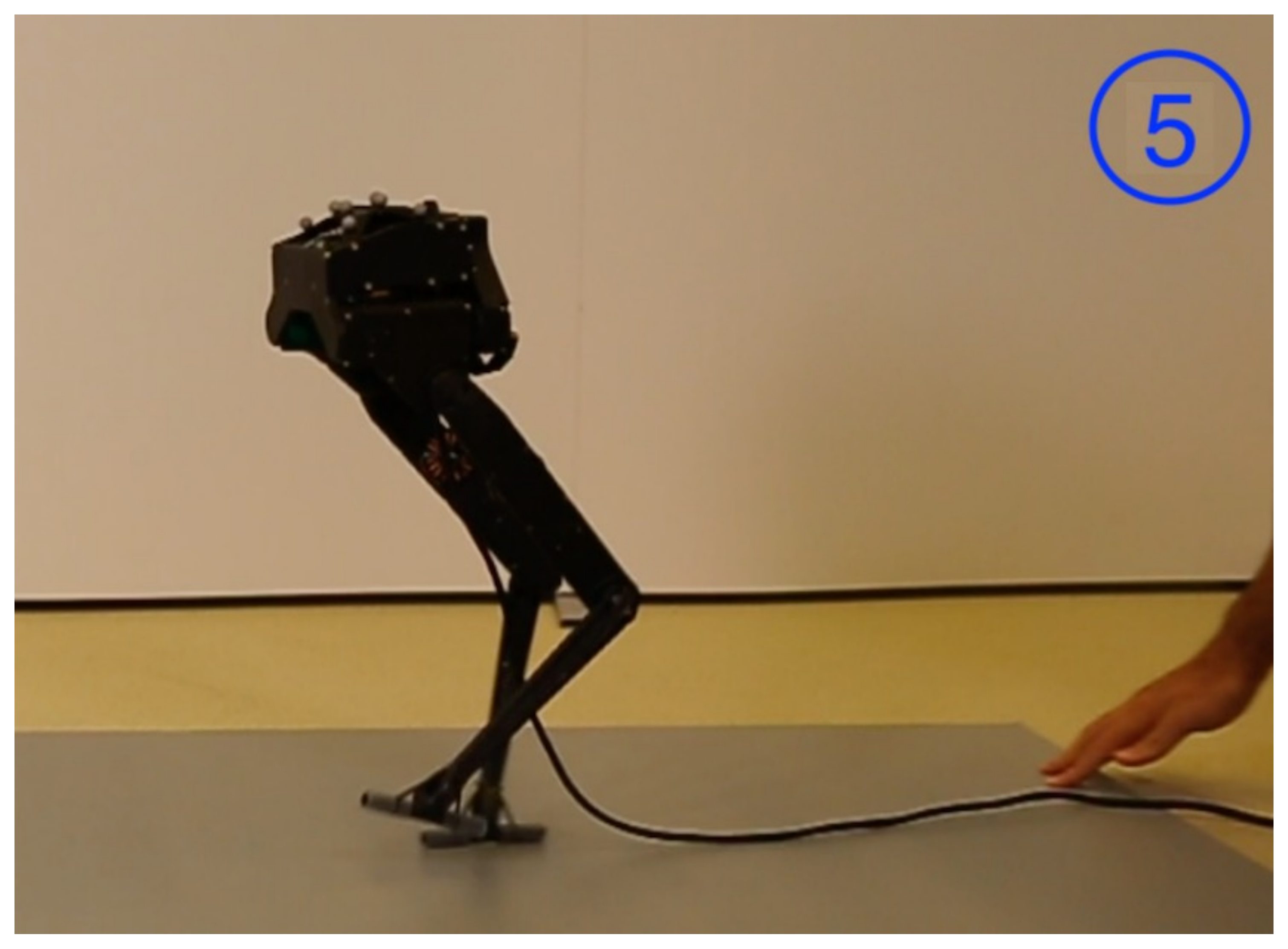}%
	\caption[]{\small Snapshots of the experiments, from left to right: 1) forward walking, 2) uneven surface, 3) soft surface, 4) push recovery, 5) slippage recovery.}
	\label{fig:snapshots}
	\vspace{-3mm}
\end{figure*}

We considered 5 different scenarios for our experiments that we will explain briefly in the following. Note that we use the exact same controller parameters for all these experiments. The snapshots of the experiments are shown in Fig. \ref{fig:snapshots}. The  accompanying video illustrates the experiments.

\subsubsection{Forward/Backward walking}\label{subsubsec:result_walking}
The robot is commanded to walk forward and backward on a flat surface with a desired velocity. Through this test we showed that our feedback control based on simplified models is robust enough to stabilize walking motion for a long time. As it can be seen in the \href{https://www.youtube.com/watch?v=x2jYQdjT_es}{video}, the robot is able to walk forward and backward for 10 minutes without any problem.

\subsubsection{Walking on uneven surfaces}\label{subsubsec:result_uneven}
We commanded stepping in place (zero desired velocity) on an uneven surface where the surface height is varied up to $\SI{4}{cm}$. The walking controller does not have any notion of the height variations; however, thanks to the adaptation of the step location and time and compliance due to torque-control, the robot can successfully step on surfaces with unknown heights.

\subsubsection{Walking on soft surfaces}\label{subsubsec:result_soft}Bolt performs in-place stepping on a soft surface (sponge, see the \href{https://www.youtube.com/watch?v=x2jYQdjT_es}{video}). This surface is specially challenging, as the stance foot does not remain fixed but oscillates due to the surface compliance. These oscillations would magnify shaking of the robot due to the structural flexibility. Although the robot would start shaking at some instances of the motion, the controller was able to damp them and the robot could successfully step on this surface without any instability.

\subsubsection{Push recovery}\label{subsubsec:result_push} In this scenario, we exert external forces to the robot's pelvis in different direction, at different times of a step and in different directions. Thanks to the fast update of the control loop, the robot quickly reacts to the pushes and adapts both step location and timing to recover. We performed push recovery scenarios both in sagittal and lateral directions.

\subsubsection{Slippage recovery}\label{subsubsec:result_slip}In the last scenario, the robot performs stepping on a surface that can slide on the ground. We disturbed this surface in different directions randomly during stepping, and the robot again was able to adapt the landing location and time of the swing foot to preserve its balance.

\begin{figure}[ht]
\centering
  \includegraphics[width=1.\linewidth]{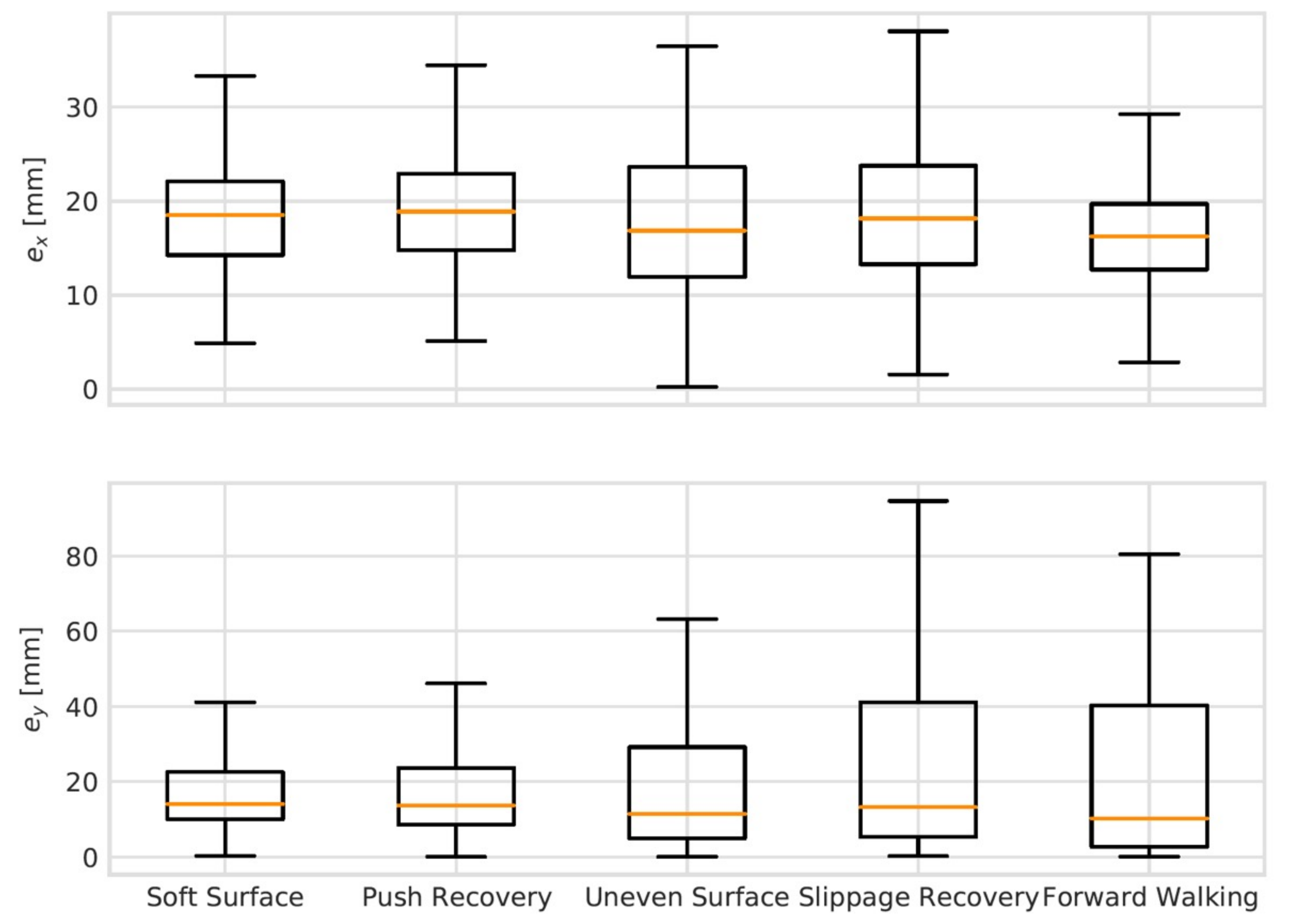}
  \caption{Box plot of the error between the desired and measured step locations for all the walking steps of the tests for both frontal and lateral directions}
  \label{fig:experiment_analysis}
  \vspace{-3mm}
\end{figure}

Finally, in Fig. \ref{fig:experiment_analysis}, we show the statistics of the error between the desired and measured landing locations for all the walking steps of the experiments through a box plot for both frontal and lateral directions. Note that since the data were skewed (i.e. they are far from being normally distributed), we used a box and whisker plot. As we can see in the figure, the medians for all the tests in both directions are below 20 mm which shows the good performance of our controller (the robot leg length in fully stretched configuration is roughly 450 mm). 

\section{Conclusions and future work}\label{sec:conclusions}
In this paper, we presented a two-level variable horizon MPC to control bipedal walking. In the high level, given the current state of the CoM, the landing location and time of the swing foot is decided. In the the lower level, the swing foot trajectory is adapted such that the swing foot lands on the ground at the desired time, as close as possible to the desired location. We integrated this two-level VH-MPC with an impedance-based WBC to compute torque commands to the robot. We conducted an extensive set of simulations and experiments through which we demonstrated the capabilities and robustness of our proposed control framework.\\

In future work, we plan to generalize our proposed approach to more general motions, i.e. 3D walking and running on more challenging terrains (e.g. stepping stones with different heights). We propose two different approaches to do this. The first approach is to perform a wider range of motions including walking and running on different surfaces in simulation and estimate the swing foot dynamics using the same approach we used in this paper. An alternative approach is to use the step location and time from the first stage and use a whole body MPC problem \cite{yeganegi2020robust} with the given contact properties as the terminal constraint/cost.\\
Extending our framework by adding collision avoidance to the swing foot MPC problem is another interesting direction to investigate. Since we are solving an MPC with the position of the foot as decision variable in the QP, collision avoidance can be dealt with using inequality constraints on the states at all times in the horizon.

Another interesting extension of the current work is alternating between the two levels of the MPC such that they come to a consensus in terms of contact location and timing at each control cycle. Finally, our approach with contact switch as a terminal constraint can be easily generalized to an event-based controller where the switch in the gait phase is triggered at the detection of swing foot contact.

\addtolength{\textheight}{0cm}   

\bibliography{Master}
\bibliographystyle{IEEEtran}

\end{document}